\pdfoutput=1

\documentclass[11pt]{article}

\usepackage{ACL2023}

\usepackage{times}
\usepackage{latexsym}
\usepackage{booktabs}
\usepackage{graphicx}
\usepackage{multirow}
\usepackage[symbol]{footmisc}

\usepackage[normalem]{ulem}
\useunder{\uline}{\ul}{}
\usepackage[colorinlistoftodos]{todonotes}
\usepackage[T1]{fontenc}

\usepackage[utf8]{inputenc}

\usepackage{microtype}

\usepackage{inconsolata}

\title{Masakhane-Afrisenti at SemEval-2023 Task 12: Sentiment Analysis using Afro-centric Language Models and Adapters for Low-resource African Languages }

\author{\normalsize  Israel Abebe Azime$^{1,\dagger}$, Sana Sabah Al-Azzawi$^{2,\dagger}$, Atnafu Lambebo Tonja$^{3,\dagger}$,\\ 
\textbf{\normalsize Iyanuoluwa Shode$^{4}$,  Jesujoba Alabi $^{1}$, Ayodele Awokoya $^{5}$, Mardiyyah Oduwole, }\\
\textbf{\normalsize  Tosin Adewumi $^{2}$, Samuel Fanijo  $^{7}$, Oyinkansola Awosan, Oreen Yousuf} \\
\footnotesize
$^\forall$Masakhane NLP, $^1$Saarland University, Germany, $^2$Luleå University of Technology, Sweden, $^3$Instituto Politécnico Nacional, Mexico,
  \\
 \footnotesize
$^4$Montclair State University, USA, $^5$University of Ibadan, Nigeria, $^6$ Brandeis University, USA, $^{7}$Iowa State University, USA.
 \\
}

\begin{document}

\maketitle
\makeatletter
\def\@fnsymbol#1{\ensuremath{\ifcase#1\or\forall\or\dagger\fi}}
\makeatother

\renewcommand{\thefootnote}{\fnsymbol{footnote}}
\footnotetext[2]{These authors contributed equally to this work.}
\renewcommand{\thefootnote}{\arabic{footnote}}

\begin{abstract}
 
In this paper, we describe our submission for the AfriSenti-SemEval Shared Task 12 of SemEval-2023. 
The task aims to perform monolingual sentiment classification (sub-task A) for 12 African languages,  multilingual sentiment classification (sub-task B), and zero-shot sentiment classification (task C).
For sub-task A, we conducted experiments using classical machine learning classifiers, Afro-centric language models, and language-specific models. For task B, we fine-tuned multilingual pre-trained language models that support many of the languages in the task. For task C, we used we make use of a parameter-efficient Adapter approach that leverages monolingual texts in the target language for effective zero-shot transfer. Our findings suggest that using pre-trained Afro-centric language models improves performance for low-resource African languages. We also ran experiments using adapters for zero-shot tasks, and the results suggest that we can obtain promising results by using adapters with a limited amount of resources.     

\end{abstract}

\section{Introduction}

Low-resourced languages receive less attention in natural language processing (NLP) research because they lack the quality datasets necessary for training, evaluation, and model implementation. However, the increasing abundance of social media platforms on the internet is changing the game for these languages, giving them more visibility and accessibility. This was not always the case, as a significant amount of NLP research focused on sentiment analysis and other tasks geared toward specific languages having a high online presence, thus resulting in the creation of techniques and models tailored to their needs \cite{yimam2020exploring}.

The rise of region-specific competitions such as the AfriSenti-SemEval competition has led to efforts to curate quality datasets sourced from the internet, create new techniques to maximize the use of these datasets for NLP tasks, and investigate the adequacy of existing NLP techniques to cater to the linguistic needs of low-resourced languages. The AfriSenti-SemEval Shared Task 12 provides a corpus of Twitter datasets written in 14 African languages for sentiment analysis tasks \cite{froebe:2023a}. This shared task presents a unique opportunity to advance sentiment analysis development in local languages and help bridge the digital divide in this area. The hope is that this contribution will assist future research and developments in this field.

This paper details our submission for the Afrisenti SemEval-2023 Task 12, which investigates the effectiveness of pre-trained Afro-centric language models and adapters for sentiment classification tasks. Our work explores the potential of these models and adapters to enhance sentiment analysis performance in the context of African languages and cultures. Our codes and all project materials are publicly available \footnote{https://github.com/SanaNGU/Sentiment-Analysis-for-African-Languages}.

\section{Background} 

Sentiment analysis, a crucial task in natural language processing employs machine learning techniques to identify emotions in text, thus having numerous practical applications in areas such as public health, business, governance, psychology, and more \cite{muhammad2022naijasenti}. The benefits of sentiment analysis are diverse with its effect felt on almost every aspect of human endeavor \cite{shode2022yosm}.

While low-resource languages have been neglected in sentiment analysis research, previous works by \citet{yimam2020exploring}, \citet{shode2022yosm} and  \citet{muhammad2022naijasenti} have created different sentiment corpora for Nigerian and  Ethiopian languages. With text classification used for benchmarking most pre-trained language models, such as \citet{ogueji-etal-2021-small} and \citet{alabi-etal-2022-adapting}, their effectiveness on the downstream task of sentiment classification is yet to be properly explored.

The dataset introduced by \citet{muhammad2023afrisenti} in SemEval 2023 Task 12, the first Afro-centric SemEval shared task, includes 14 African languages divided into three sub-tasks: language-specific, multilingual, and zero-shot competitions. As part of our contribution, our team participated in all sub-tasks of the shared task, explored the effectiveness of several pre-trained networks trained on African languages, and examined the use of parameter-efficient approaches like adapters \cite{pfeiffer2020mad} for a zero-shot classification task. 

\section{System Overview}
\subsection{Development Phase}
In the development phase of the competition, we tested several algorithms. First, we worked with unigram count of words and Tf-idf normalized word count features. These features use very simplified word-counting strategies. Using these features, we tested Multinomial Naive Bayes, Multi-Layer Perception, and XGB classifiers. 

The next phase of our experiment focuses on pre-trained language models. We worked on Afro-XLMR (small, base) by \citet{alabi-etal-2022-adapting}, LaBSE by \citet{feng2020languageagnostic}, AfriBERTa-Large by \citet{ogueji-etal-2021-small}, AfroLM by \citet{dossou2022afrolm}, Bernice by \citet{delucia2022bernice}, QARiB by \citet{abdelali2021pretraining}, and other language and task-specific models, such as \citet{DBLP:journals/corr/abs-2104-12250},  DarijaBERT\footnote{https://huggingface.co/SI2M-Lab/DarijaBERT} and 
twitter-xlm-roberta-base-sentiment\footnote{https://huggingface.co/cardiffnlp/twitter-xlm-roberta-base-sentiment}. We further experimented with the text-to-text work done by \citet{jude-ogundepo-etal-2022-afriteva}, and on adapters, as proposed by \citet{pfeiffer2020adapterhub}.
We found some unusual predictions when experimenting with mT5-based \cite{xue-etal-2021-mt5} Afro-centric models, which was observed in previous work as well \cite{adewumi-etal-2022-ml,adewumi2022t5}.

During the development phase of both language-specific Task A and multilingual Task B, we trained the previously mentioned models. We also performed data cleaning by removing '@user' tags and eliminating punctuation and emojis. However, we observed no significant improvement after this cleaning process, so we conducted further experiments without cleaning the dataset and used it in its original state.

For Task C, we investigated the linguistic similarities between Tigrinya, Oromo, and other languages. To aid in this exploration, we drew from the previous work of \citet{10.1007/978-3-319-95153-9_13}, who examined the similarities between Amharic and Tigrinya. Leveraging the Amharic dataset we had, we also worked on translating it to Tigrinya using Meta's No Language Left Behind (NLLB) \cite{costa2022no} MT model that promises to deliver high-quality translations across 200 languages.

When working with the Oromo language, we were initially uncertain about its linguistic similarities with other languages. To address this, we experimented with using a language spoken in the same region and other languages within the same family, such as Hausa. After conducting our experiments, we found that training models on Hausa data and then using these same models to predict outcomes in Oromo produced better results than with other languages. 

For the zero-shot cross-lingual tasks, we used adapters \cite{pfeiffer2020adapterhub} while following the two-step procedure as proposed by \citet{pfeiffer2020mad}.
The procedure is as follows: \textbf{(i)} trained language-specific adapters using monolingual data, and \textbf{(ii)} trained task adapters using the task-specific dataset.
For the first step, we trained language-specific adapters for Tigrinya and Oromo using a monolingual dataset and AfroXLMR-base \cite{alabi-etal-2022-adapting} as our base model.
Since the two languages are similar, we used the Amharic training dataset for the task adapter for the Tigrinya zero-shot task, as described in the previous paragraph. Hausa and Swahili datasets were used to train the task adapters for the Oromo zero-shot task as was not a similar language to Oromo in the given training dataset.

We used language adapters for Amharic, Hausa, and Swahili from \citet{david_ifeoluwa_adelani_2022_7416488}. We trained the Amharic task adapter by using the Amharic language adapter and an Amharic sentiment dataset. After training the Amharic task adapter we replaced the Amharic language adapter with the Tigrinya language adapter and evaluated the zero-shot performance for Tigrinya. Similarly, for Oromo, we trained Swahili and Hausa task adapters using Swahili and Hausa language adapters, respectively. After training task adapters, we replaced both language adapters with Oromo and tested their performance for the Oromo zero-shot task.

\subsection{Test Phase}\label{test phase}

In the competition's Task A testing phase, we selected the top three models from the ones we trained during the development phase. After identifying the better-performing models for each language, we submitted our results based on those models. For our fourth submission, we used a voting-based ensemble approach with the previous three models. Finally, for our last submission, we used the best-performing multilingual model trained on all available training and validation data to predict a specific language.

In multilingual Task B, we identified the four best-performing models and submitted four entries using them. Based on the models' observed performance, we employed a voting ensemble approach with the top three models for our final submission. For Task C, we submitted five predictions based on the model's performance during the development phase.

\section{Experimental Setup}

\subsection{Dataset and Evaluation Metrics}

\begin{figure*}[h!]
\centering
\includegraphics[width=0.75\textwidth]{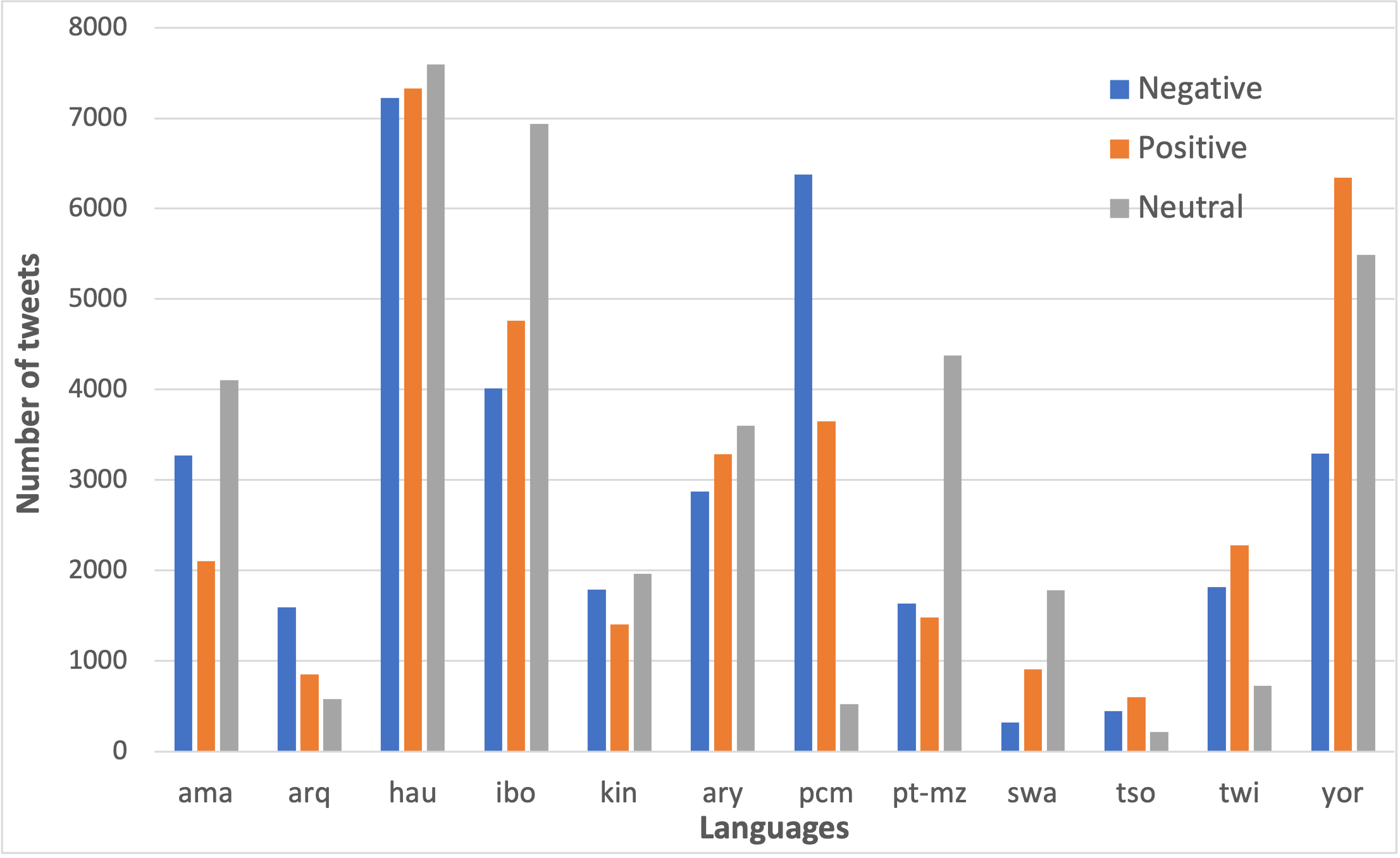}
\caption{Data distribution among the languages.}
\label{Figure 1}
\end{figure*}

During the computation and development phases, we were given only training data with labels and development data without labels. The training dataset was originally from the AfriSenti dataset, which is a corpus of 14 African languages scraped from Twitter for sentiment analysis tasks. To perform experiments repeatedly on the dataset, we created our own training and evaluation data for development, treating the provided development data as a test set. This enabled us to test a list of models we had planned to experiment with. Due to the computationally intensive nature of training, we filtered the best models based on the created data. The promising models based on this data were used in the test phase. We utilized a weighted F1 score as the evaluation metric for our models.


As shown in Figure \ref{Figure 1}, the dataset across all languages is imbalanced. This problem is more pronounced when the language datasets are joined to form multilingual data. To resolve this, during our development phase, we attempted label-based and language-based balancing, which involved performing repetitive sampling on data with low frequency. However, these experiments did not yield any significant improvement, therefore we did not include them in our final training for submission. 

\subsection{Training}

To conduct experiments using pre-trained BERT models, we utilized the Huggingface\footnote{https://huggingface.co/} and PyTorch library\footnote{https://pytorch.org/}. Our implementation of the sentiment analysis models was based on code from the Afrisenti-SemEval GitHub repository\footnote{https://github.com/afrisenti-semeval/afrisent-semeval-2023}. We made modifications to the code to enable repeated training for all models and performed data processing specific to the provided data. Additionally, we explored the use of text-to-text models from the Afriteva GitHub repository\footnote{https://github.com/castorini/afriteva} and employed adapter-hub implementations\footnote{https://github.com/adapter-hub/adapter-transformers} for our adapter-related experiments.

We used existing hyper-parameters for most of our experiments, except for the pre-trained language models which we trained for 10 epochs during the testing phase. We selected the weights from the best epoch with the highest evaluation F1 score for submission.

For our final submissions, we combined as much data as possible for training. In task A, we used language-specific training and gold label data, while in task B, we utilized all available training and development datasets. Unfortunately, our experiment aimed at incorporating language-specific tags was unsuccessful, as the dataset did not provide the necessary language tags.
\begin{table*}[h]
\resizebox{\textwidth}{!}{
\begin{tabular}{|l|llllllllllll|}
\hline
\multirow{2}{*}{\textbf{Submission}} & \multicolumn{12}{c|}{\textbf{Language}} \\ \cline{2-13} 
& \multicolumn{1}{l|}{ama} & \multicolumn{1}{l|}{ary}  & \multicolumn{1}{l|}{hau}   & \multicolumn{1}{l|}{ibo}   & \multicolumn{1}{l|}{yor}  & \multicolumn{1}{l|}{twi}   & \multicolumn{1}{l|}{pcm} & \multicolumn{1}{l|}{arq}     & \multicolumn{1}{l|}{pt-mz}     & \multicolumn{1}{l|}{swa}     & \multicolumn{1}{l|}{kin}     & tso     \\ \hline

1 & \multicolumn{1}{l|}{60.60\textsuperscript{1}} & \multicolumn{1}{l|}{\textbf{61.27\textsuperscript{2}}} & \multicolumn{1}{l|}{77.97\textsuperscript{3}} & \multicolumn{1}{l|}{77.26\textsuperscript{4}} & \multicolumn{1}{l|}{73.45\textsuperscript{3}} & \multicolumn{1}{l|}{63.76\textsuperscript{3}} & \multicolumn{1}{l|}{66.11\textsuperscript{3}} & \multicolumn{1}{l|}{70.34\textsuperscript{6}} & \multicolumn{1}{l|}{67.63\textsuperscript{3}} & \multicolumn{1}{l|}{58.80\textsuperscript{3}} & \multicolumn{1}{l|}{65.80\textsuperscript{5}} & 50.48\textsuperscript{7} \\

2 & \multicolumn{1}{l|}{51.49\textsuperscript{3}} & \multicolumn{1}{l|}{57.29\textsuperscript{8}} & \multicolumn{1}{l|}{77.57\textsuperscript{1}} & \multicolumn{1}{l|}{77.26\textsuperscript{8}} & \multicolumn{1}{l|}{65.54\textsuperscript{8}} & \multicolumn{1}{l|}{63.56\textsuperscript{8}} & \multicolumn{1}{l|}{67.18\textsuperscript{4}} & \multicolumn{1}{l|}{70.34\textsuperscript{8}} & \multicolumn{1}{l|}{71.26\textsuperscript{3}} & \multicolumn{1}{l|}{57.88\textsuperscript{7}} & \multicolumn{1}{l|}{67.28\textsuperscript{4}} & 53.00\textsuperscript{1}  \\

3      & \multicolumn{1}{l|}{53.93\textsuperscript{4}}  & \multicolumn{1}{l|}{58.71\textsuperscript{9}}  & \multicolumn{1}{l|}{\textbf{78.53\textsuperscript{4}}} & \multicolumn{1}{l|}{\textbf{78.42\textsuperscript{1}}} & \multicolumn{1}{l|}{\textbf{73.85\textsuperscript{1}}} & \multicolumn{1}{l|}{\textbf{64.62\textsuperscript{1}}} & \multicolumn{1}{l|}{67.18\textsuperscript{10}}  & \multicolumn{1}{l|}{\textbf{72.86\textsuperscript{2}}} & \multicolumn{1}{l|}{\textbf{71.57\textsuperscript{8}}} & \multicolumn{1}{l|}{\textbf{59.95\textsuperscript{4}}} & \multicolumn{1}{l|}{\textbf{67.31\textsuperscript{3}}} & 52.00\textsuperscript{5} \\

4 (Ensemble)    & \multicolumn{1}{l|}{60.60}  & \multicolumn{1}{l|}{61.07}  & \multicolumn{1}{l|}{77.57}  & \multicolumn{1}{l|}{77.26}  & \multicolumn{1}{l|}{65.54}  & \multicolumn{1}{l|}{63.56}  & \multicolumn{1}{l|}{\textbf{67.18}} & \multicolumn{1}{l|}{70.34}  & \multicolumn{1}{l|}{67.63}  & \multicolumn{1}{l|}{58.80}  & \multicolumn{1}{l|}{65.80}  & 52.00  \\

5 (LaBSE on All Data) & \multicolumn{1}{l|}{\textbf{68.85}} & \multicolumn{1}{l|}{55.50}  & \multicolumn{1}{l|}{73.12}  & \multicolumn{1}{l|}{73.75}  & \multicolumn{1}{l|}{68.98}  & \multicolumn{1}{l|}{62.83}  & \multicolumn{1}{l|}{66.88}  & \multicolumn{1}{l|}{64.35}  & \multicolumn{1}{l|}{70.26}  & \multicolumn{1}{l|}{60.24}  & \multicolumn{1}{l|}{64.14}  & \textbf{54.33}  \\\hline
\end{tabular} \\
}

\caption{\label{taska}  Test phase comparison of F1-scores (\%) for Task A sentiment analysis on 12 African languages using various pre-trained  models. Models include (1) AfriBerta-large, (2) Dziribert, (3) LaBSE, (4) AfroXLMR-base, (5) AfroXLMR-small, (6) DarijaBERT, (7) AfroLM, (8) twitter-xlm-roberta-base-sentiment, (9) QARiB , and (10) Bernice.}
\end{table*}

\section{Results}

\subsection{Task A}
Our final submission for each of the monolingual datasets was the result of LaBSE multilingual model on each language as described in \ref{test phase}. Although this model did not give us the best F1 score for all languages, it worked well for Amharic and Xitsonga - resulting in our team ranking 6th out of 29, and 8th out of 31, respectively, on the leaderboard. DziriBERT produced the best F1 score for Moroccan Arabic/Darija and Algerian Arabic, thus ranking our team 14th out of 32, and 19th out of 30 participants, respectively.

Afro-XLMR-base gave rise to our best result for Hausa and Swahili languages with our team ranking 31st out of 35 and 16th out of 30 participants. AfriBerta-Large produced the best F1 score for Igbo, Yoruba, and Twi languages resulting in our team ranking 26th out of 32, 23rd out of 33, and 21st out of 31 for each language, respectively. The ensemble approach on LaBSE, Afro-XLMR-base, and Bernice model produced the best prediction for the Nigerian Pidgin language, resulting in our team ranking 16th out of 32. Twitter-XLM-Roberta and LaBSE models predicted the best F1 scores for Mozambican Portuguese and Kinyarwanda, resulting in our team ranking 13th out of 30, and 20th out of 34, respectively. The overview of the F1 scores for each of the models we considered can be found in Table
\ref{taska}.

\subsection{Task B}

Our best model was an ensembled AfroXLMR-base, LaBSE multilingual, and twitter-xml-roberta-base-sentiment models for the multilingual sentiment classification task. This ensemble model put our team in the 9th position out of 33. The models considered and their F1 scores for this task are described in table \ref{taskb}. 

\begin{table}[h]
\resizebox{
\columnwidth}{!}{
\begin{tabular}{|l|l|l|}
\hline
\textbf{S.No.} & \textbf{Model}    & \textbf{F1-Score (\%)} \\ \hline
1  & AfroXLMR-base     & 67.69   \\ \hline
2   & AfroLM     & 62.45    \\ \hline
3    & LaBSE\_multilingual     & 68.28  \\ \hline
4   & twitter-xlm-roberta-base-sentiment    & 66.07 \\ \hline
5  & Ensamble of Top 3& \textbf{70.34}  \\ \hline
\end{tabular}
}
\caption{\label{taskb} Test phase comparison of F1-scores (\%)  for Task B}
\end{table}
\subsection{Task C}

For the zero-shot classification on Tigrinya (Table \ref{taskc1}), our last submitted model was an ensembled AfroXLMR trained on Amharic, multilingual AfroXLMR, and multilingual AfriBerta. This placed our team 18th out of 28. However, our best model was multilingual AfroXLMR which produced an F1 score of 61.48\%, as opposed to our submitted ensembled model which produced an F1 score of 57.99\%. For the same task on the Oromo language, our ensembled AfroXMLR trained on the multilingual dataset, AfriBerta trained on the multilingual dataset and the adapter model produced the best F1 score during our training process. This resulted in our team ranking 10th out of 29. 

\begin{table}[h]
\resizebox{\columnwidth}{!}{
\begin{tabular}{|l|l|l|}
\hline
\textbf{S.No.} & \textbf{Model}  & \textbf{F1-Score (\%)} \\ \hline
1 & Adapter   & 32.95  \\ \hline
2  & AfroXLMR (Amharic all+ tig dev)   & 57.99  \\ \hline
3    & AfroXLMR (Multilingual + tig dev) & \textbf{61.48}   \\ \hline
4    & AfriBerta (Multilingual + tig dev)  & 54.74  \\ \hline
5     & Top 3 Ensemble       & 57.99    \\ \hline
\end{tabular}
}

\caption{\label{taskc1} Test phase comparison of F1-scores (\%)  for Task C-Tigrinya}
\end{table}

\begin{table}[h]
\resizebox{\columnwidth}{!}{
\begin{tabular}{|l|l|l|}
\hline
\textbf{S.No.} & \textbf{Model}                                                                    & \textbf{F1-Score (\%)} \\ \hline
1                           & AfroXLMR Multilingual data + orm dev ) & 36.89                         \\ \hline
2                           & AfriBerta (Multilingual data + orm dev )& 42.09                         \\ \hline
3                           & Adapter & 34.08                         \\ \hline
4 & Top 3 Ensemble   & \textbf{42.09}                \\ \hline
\end{tabular}
}
\caption{\label{taskc2} Test phase comparison of F1-scores (\%) for Task C-Oromo}
\end{table}

	
	
	

\section{Conclusion}
In this paper, we presented our submission for the AfriSenti-SemEval Shared Task 12 of SemEval-2023, focusing on the classification of sentiment for monolingual, multilingual, and zero-shot settings for low-resource African languages. We explored Afro-centric, language-specific, and general pre-trained language models for fine-tuning. Based on the test result, for monolingual sentiment classification, Afro-centric language models show better performance for most of the languages. However, language-specific pre-trained language models perform better than Afro-centric language models for Algerian Arabic and Morrocan Darija. For multilingual sentiment classification, Afro-centric language models showed promising results. For the zero-shot task, we see that using adapters shows promising results for related languages while Afro-centric languages show better performance for unrelated languages.              

We would like to further our research on adapters for low-resource languages as it shows promising results in the zero-shot setting for related languages.


\bibliography{semeval23-clickbait-spoiling-system-paper}
\bibliographystyle{acl_natbib}



\end{document}